\documentclass[runningheads]{llncs}

\usepackage{eccv}

\usepackage{eccvabbrv}
\usepackage{multirow}
\usepackage[table]{xcolor} 
\usepackage{booktabs}
\usepackage{amssymb}
\usepackage{bbm}
\usepackage[accsupp]{axessibility}  

\usepackage{hyperref}

\usepackage{orcidlink}
\usepackage{comment}

\begin{document}

\title{VisLens: Single-Pass Interpretable Visual Search for Multimodal LLMs} 





\author{Jingyi He\inst{1} \and
Sanghwan Kim\inst{1,2} \and
Zeynep Akata\inst{1,2}}
\authorrunning{J.~He et al.}
\institute{Technical University of Munich \quad Helmholtz Munich \and
Munich Center for Machine Learning
}

\maketitle

\begin{abstract}
Multimodal large language models (MLLMs) struggle with fine-grained Visual Search, the task of locating small or rare objects in high-resolution images. Existing remedies fall into two families: (1) Training-free methods based on attention or confidence scores are accurate but slow, since they require multiple MLLM queries per example. (2) Reinforcement Learning (RL) trained tool-use models are faster at inference but opaque, since their tool calls remain uncontrollable and hard to interpret. To overcome this, we propose \emph{VisLens} (Visual Focus via Logit Lens), a Visual Search method built on the logit lens, which decodes the semantics held in a hidden state by projecting it through the LLM head. VisLens further uses a lightweight tuned-lens that maps early hidden states into the final hidden state space, so visual tokens can be read out from early layers. These tokens are matched to target words in the query to generate a crop of the relevant region, which is fed back in alongside the original image to produce the final answer. The whole process, from decoding to the final answer, completes in a single forward pass without repeated queries. VisLens matches or exceeds prior baselines while delivering a
substantial latency advantage, running $8.5$--$9.9\times$ faster than Thyme
and up to $22.2\times$ faster than training-free multi-pass search methods.

\keywords{Multimodal LLM \and Visual Search \and Interpretability}
\end{abstract}

\section{Introduction}

Multimodal large language models (MLLMs) perform well on coarse vision and language tasks such as captioning and general visual question answering, but they struggle with fine-grained Visual Search, where the target occupies only a tiny fraction of a high-resolution image~\cite{wu2024v,wang2025divide,zhang2025mllms}. The target signal is easily lost among irrelevant background content, and the model's global representation is often too coarse for precise localization. This gap matters for applications such as document analysis, surveillance, and autonomous driving, where finding small details reliably is essential~\cite{zhang2024mme}.

Existing remedies fall into two families, each with a notable cost. The first is training-free: methods such as ViCrop~\cite{zhang2025mllms}, FOCUS~\cite{zhong2026focus}, ZoomEye~\cite{shen2024zoomeye}, and UG-Search~\cite{kim2025training} inspect attention patterns or uncertainty scores to find candidate regions, zoom in, and query the model again to refine the result. This requires multiple MLLM queries per example, which multiplies inference cost and limits practicality in time sensitive settings. The second family trains models to use tools, such as cropping or zooming functions, through reinforcement learning. Methods such as DeepEyes~\cite{zheng2025deepeyes}, Thyme~\cite{zhang2025thyme}, and Mini-o3~\cite{lai2025mini} learn this tool-use behavior from specialized instruction data. While accurate, this requires costly training pipelines built around task specific datasets, and the resulting policy is a black box: it produces a crop region that is not fully controllable or interpretable, and it does not always explain why that region was chosen, which makes failures hard to diagnose.

Our key insight is that an MLLM does not need to be queried repeatedly or trained to use tools in order to know where relevant content lies. Early visual-token states of the MLLM already encode semantic information about the content at different spatial locations in the image. If this information is decoded from a translated early visual-token state and matched against the words describing the search target, the target's location can be recovered directly, and the corresponding crop can be fed back in alongside the original image within the same forward pass, without extra queries or reinforcement learning.

We operationalize this insight in \emph{VisLens} (Visual Focus via Logit Lens), a method built on the logit lens~\cite{nostalgebraist2020logitlens,wang2022interpretability,wendler2024llamas,wang2025logitlens4llms}, a technique widely used to interpret the semantic space of LLMs and recently shown to work similarly well for visual tokens in MLLMs~\cite{phukan2025beyond,jiang2025interpreting,neo2025towards}. The logit lens projects a hidden state at any layer through the model's output head to approximate what the model would predict at that point. VisLens uses this projection to decode the semantics of each image region, match the decoded tokens against target words from the query, and merge the best matching regions into crop boxes through clustering. These boxes are then fed alongside the original image to produce the final answer. Since the crop comes from an early MLLM layer, the whole process fits within a single forward pass, avoiding the latency of repeated queries while staying transparent, as each localization decision traces back to specific decoded tokens.

Applying the logit lens directly to early layers is unreliable, since the output head is tuned to final layer representations rather than the partially formed ones seen earlier in the network. To address this, VisLens trains a lightweight tuned-lens~\cite{belrose2023eliciting}, a small MLP that maps early hidden states into the final layer's representational space before decoding. This mapping is cheap to train, requires no change to the underlying MLLM, and substantially improves early layer localization. VisLens matches or
outperforms prior visual-search baselines while delivering a substantial latency
advantage, running $8.5$--$9.9\times$ faster than Thyme\cite{zhang2025thyme} and up to
$22.2\times$ faster than training-free multi-pass search methods.
In summary, our contributions are as follows:

\begin{enumerate}
  \item We propose VisLens, a single-pass and interpretable method for Visual Search based on the logit and tuned-lens.
  \item The VisLens pipeline is fully controllable and interpretable, turning decoded tokens into crop regions through semantic matching, clustering, and box merging.
  \item We observe a favorable accuracy and latency trade-off across V$^*$Bench and HR-Bench, supported by interpretability case studies.
\end{enumerate}

\section{Related Work}

\noindent\textbf{Visual Search in MLLMs.} Visual Search asks an MLLM to locate a small or rare target in a high-resolution image, a task formalized by V$^*$Bench and its proposed SEAL framework~\cite{wu2024v}. Existing solutions fall into either training-free or RL-tuned methods. First, training-free approaches use attention or confidence signals to propose and zoom into candidate regions, including ViCrop~\cite{zhang2025mllms}, \mbox{FOCUS}~\cite{zhong2026focus}, ZoomEye~\cite{shen2024zoomeye}, and UG-Search~\cite{kim2025training}. These need no extra training, but each candidate region must be re-encoded and re-queried, costing several forward passes per example. Second, RL-tuned approaches train an MLLM to invoke cropping or zooming tools, as in DeepEyes~\cite{zheng2025deepeyes}, Thyme~\cite{zhang2025thyme}, and Mini-o3~\cite{lai2025mini}. These avoid repeated queries at test time but require costly tool-use instruction data and give no explicit account of why a region was chosen. VisLens combines the training-free efficiency with a single forward pass, while offering the interpretability neither approach provides.

\noindent\textbf{Logit lens for MLLMs.} The logit lens reads what LLMs believe at intermediate layers by projecting the hidden state through the final output head~\cite{nostalgebraist2020logitlens}. This simple technique has been used to analyze LLM circuits~\cite{wang2022interpretability}, multilingual representations~\cite{wendler2024llamas}, and modern architectures more broadly~\cite{wang2025logitlens4llms}. Since the output head is tuned only to final layer representations, the raw lens is unreliable early on. Then, the tuned-lens fixes this with a lightweight per-layer mapping into the final layer space~\cite{belrose2023eliciting}. More recently it has been extended to MLLMs to detect hallucinations~\cite{phukan2025beyond,jiang2025interpreting} and to show that decoded visual tokens localize to the true object position~\cite{neo2025towards}. In all of this work, however, the lens serves purely as a diagnostic that explains behavior after the fact. VisLens instead turns the logit and tuned-lens into the mechanism that drives single-pass Visual Search itself.

\section{Visual Focus via Logit Lens (VisLens)}

\noindent \textbf{Task definition.}
Visual Search takes a high-resolution image $I$ and a natural-language query $q$
referring to a target object, and requires the model to output the answer
$y = \operatorname{MLLM}(I, q)$. 

The defining difficulty is that the target occupies
only a tiny fraction of the scene, while MLLMs encode the image under a fixed
resolution and token budget, so the target's signal is diluted below the model's
effective perceptual threshold. Answering thus decomposes into locating the small
target region, a crop $I_c$ of $I$, and then reasoning over it, so accuracy is bottlenecked by localization rather than by the reasoning that
follows, a gap that correlates with target size. VisLens therefore recovers
a crop $I_c$ approximating $I_c^\star$ from $I$ and $q$ alone.

\noindent \textbf{Logit lens preliminaries.}
The MLLM encodes the high-resolution image $I$ into $N$ visual tokens on a grid,
each token $i$ carrying a hidden state $h_\ell^{(i)} \in \mathbb{R}^{d}$ at layer $\ell$.
The logit lens reads out intermediate computation by applying the frozen LM head, namely the final layer norm followed by the unembedding $W_U$, to the hidden state,
\[
  \phi(h) = \operatorname{softmax}\!\big(W_U\,\operatorname{LN}(h)\big) \in \Delta^{|\mathcal{V}|},
\]
yielding a vocabulary distribution that approximates the prediction the model would make
if it decoded from that layer. Applied at a visual position, $\phi\big(h_\ell^{(i)}\big)$
maps the patch embedding to tokens whose top entries name the object or attribute encoded
there, exposing per-token visual semantics without training any probe. Because visual tokens
retain a one-to-one correspondence with grid cells, these per-token distributions assemble
into a spatial map that reveals where a given concept is represented in the image.

\noindent \textbf{Tuned-lens.}
The logit lens applies the frozen head $\phi$ to an intermediate residual-stream
state $h_\ell$, but its predictions deteriorate at early layers. The cause is a
representation mismatch: the unembedding inside $\phi$ is trained to decode only the
final state $h_{L}$, while later transformer blocks progressively change the basis
and scale of the residual stream. An early state may already encode the eventual
prediction, yet not in a form the final head can directly read.

We therefore insert a lightweight translator $g_\ell$ that maps a source-layer state
into the final-layer space before the head is applied,
\[
  g_\ell(h) = h + \operatorname{MLP}_\ell(h),
\]
a residual MLP with a bottleneck inside the $d$ dimension hidden
space, initialized near identity ($\operatorname{MLP}_\ell \approx 0$) so training
starts from the plain logit lens solution. The tuned-lens readout at layer $\ell$
is then
\[
  \operatorname{TL}_\ell(h) = \phi\big(g_\ell(h)\big),
\]
with a separate $g_\ell$ trained per source layer (default target: the final layer
$L$).

Training is knowledge distillation in vocabulary space. Restricting to visual-token
positions $\mathcal{P}_{\text{vis}}$, we match the translated readout to the model's
own final-layer distribution by minimizing the token-wise KL divergence,
\[
  \min_{\theta_\ell}\;
  \mathbb{E}_{(I,q)}\,\mathbb{E}_{\,i \in \mathcal{P}_{\text{vis}}}\;
  D_{\mathrm{KL}}\!\Big(
    \phi\big(h_{L}^{(i)}\big)\;\big\|\;
    \phi\big(g_\ell(h_\ell^{(i)})\big)
  \Big),
\]
where the teacher $\phi(h_{L}^{(i)})$ is the model's own final hidden state, so no
external labels are required. Unlike hidden-state regression, this objective directly
optimizes the distribution used for token decoding.

Training is lightweight. The base MLLM stays frozen and runs without gradients; only
the translator parameters $\theta_\ell$ are
updated, which account for only a small fraction of the full model ($\sim\!0.05\%$ for an $8\text{B}$ MLLM), and supervision is confined to the visual positions where the early-layer
logit lens is least reliable. The translator thus acquires no
new task knowledge, learning only the change of basis that lets the frozen head decode
early visual representations.
  
\noindent \textbf{Full pipeline.}
A single forward pass, truncated at the source layer, decodes every visual cell
through the tuned-lens into its top-$K$ tokens. Writing $\operatorname{top\text{-}}K$
for the map from a distribution to its $K$ highest-probability (token, probability)
pairs, this assembles a semantic map
\[
  \operatorname{TL} = \{S_i\}_{i=1}^{N}, \qquad
  S_i = \operatorname{top\text{-}}K\big(\operatorname{TL}_\ell(h_\ell^{(i)})\big),
\]
one top-$K$ token--probability list $S_i$ per visual cell. VisLens turns this map and
the query into target crops, $I_c = \operatorname{VisLens}(\operatorname{TL}, q)$,
unrolled as:

\begin{figure}[t]
  \centering
  \includegraphics[clip, trim=10 90 180 68, width=\linewidth]{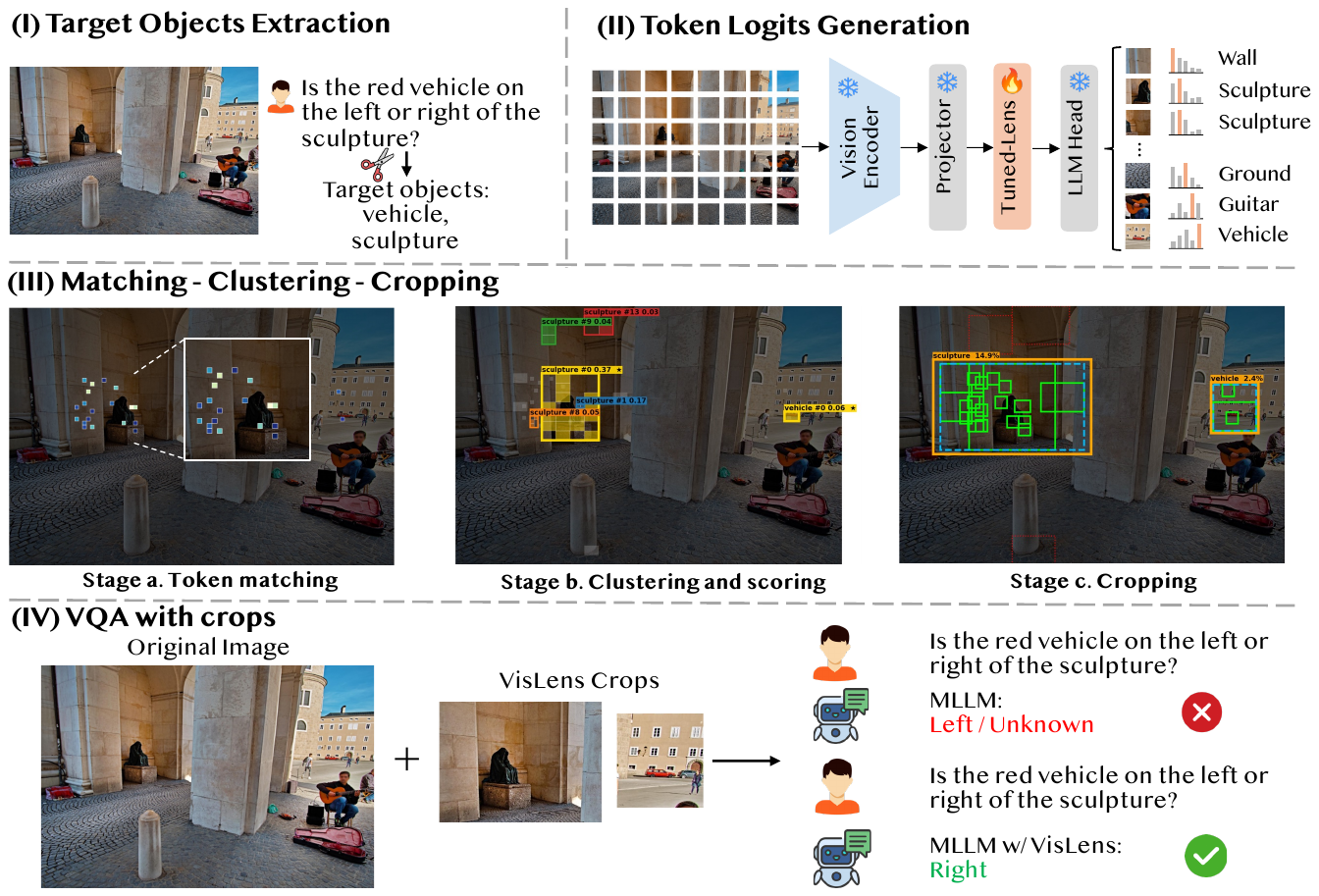}
  \caption{\textbf{VisLens pipeline.} From the tuned-lens map $\{S_i\}$, VisLens localizes each query target and
constructs crops that are fed together with the original image to the same frozen
MLLM. Only the translator is trained.}
  \label{fig:pipeline}
\end{figure}

\begin{enumerate}
\item \textbf{Target extraction.} POS-tag $q$ using WordNet\cite{miller1995wordnet} and keep content nouns that denote
      localizable objects, dropping attribute words (\emph{color}, \emph{size},
      \emph{shape}) and observer nouns (\emph{camera}, \emph{viewer}); adjacent nouns
      merge into one phrase (\emph{fire hydrant}), so each phrase names a single target:
      \[
        P = \operatorname{Extract}(q) = \{p_1,\dots,p_M\}.
      \]

  \item \textbf{Semantic matching.} Score each cell for a phrase by the probability of its
        best-matching lens token, where a phrase matches if any of its words does:
        \[
          H_j[i] = \max_{(t,\pi)\in S_i} \pi\,\mathbbm{1}\!\left[t \in p_j\right].
        \]
        This gives one probability heatmap $H_j$ per phrase. A synonym fallback
        (\emph{handbag}$\to$\emph{bag}/\emph{purse}) applies only when a phrase matches nowhere.

\item \textbf{Clustering.} Group the matched cells ($H_j[i] > 0$) into connected
      components under 8-adjacency ($\operatorname{CC}_8$), scoring each by its mass:
      \[
        \mathcal{C}_j = \operatorname{CC}_8\!\big(\{\,i : H_j[i] > 0\,\}\big), \qquad
        \operatorname{score}(C) = \sum_{i\in C} H_j[i].
      \]

\item \textbf{Box merging.} Per phrase, union the components strongest-first, stopping
      before the enclosing box would exceed a fraction $\tau$ of the image area, giving one
      capped box $B_j$ per target. VisLens then crops each target box independently,
      \[
        I_c = \{\operatorname{Crop}(I, B_j)\}_{j=1}^{M},
      \]
      producing one crop per target. As an ablation, we also consider a union
      variant that merges all target boxes into a single crop
      $\operatorname{Crop}\big(I,\ \bigcup_{j=1}^{M} B_j\big)$, preserving their spatial
      relation.
      For models that encode both high-resolution tiles and a low-resolution thumbnail
      (e.g.,~LLaVA-OneVision\cite{li2024llavaonevisioneasyvisualtask}, InternVL3\cite{zhu2025internvl3exploringadvancedtraining}), the boxes from the two grids are unioned in
      pixel space before cropping.
\end{enumerate}

\section{Experiments}

\subsection{Setup}

\noindent\textbf{Models.} We apply VisLens to several open-source MLLMs: LLaVA-OneVision\cite{li2024llavaonevisioneasyvisualtask},
Qwen2.5-VL\cite{bai2025qwen25vltechnicalreport}, and InternVL3\cite{zhu2025internvl3exploringadvancedtraining}.
The backbone is frozen throughout; VisLens adds only the tuned-lens translator and
the crop-and-reprompt loop, with no fine-tuning of the base model.

\noindent\textbf{Benchmarks.} We evaluate on visual-search benchmarks: 
V$^{*}$Bench\cite{wu2024v}, HR-Bench\cite{wang2025divide}, where targets occupy a small fraction
of high-resolution scenes, and on standard VQA benchmarks, 
A-OKVQA\cite{aokvqa}, GQA\cite{gqa}, and
POPE\cite{pope}, to verify that VisLens does not degrade performance on
tasks without a small-target bottleneck.

\noindent\textbf{Baselines.} We first compare VisLens against its unmodified backbone in Sec.~\ref{sec:main result},
isolating the effect of our method
(Tab.~\ref{tab:visual-search-qa}). We then compare against existing visual-search methods on an
accuracy--latency trade-off (Fig.~\ref{fig:acc-latency}): the training-free method
ZoomEye\cite{shen2024zoomeye}, UG-Search\cite{kim2025training}, and the RL-trained Thyme\cite{zhang2025thyme}. All methods share the same backbone where applicable, so differences
reflect the search mechanism rather than the underlying MLLM.

\noindent\textbf{Metrics.} We report task accuracy together with average wall-clock latency per query.
Accuracy alone cannot distinguish the methods: all visual-search approaches
trade extra processing for better localization, so latency is needed to
capture the cost at which each method obtains its accuracy.

\noindent\textbf{Tuned-lens training.} For each backbone we train one residual-MLP
translator per source layer, supervised only at visual-token positions by KL
distillation against the model's final-layer distribution.


\subsection{Main Results}
\label{sec:main result}

\begin{table*}[htbp]
\centering
\caption{Performance on Visual-Search QA benchmarks. Subscripts in
\textcolor[HTML]{2E8B57}{green} denote the improvement of \textbf{VisLens}
over each base model.}
\label{tab:visual-search-qa}
\small
\setlength{\tabcolsep}{4pt}
\renewcommand{\arraystretch}{0.95}
\resizebox{\textwidth}{!}{%
\begin{tabular}{l ccc ccc ccc}
\toprule
\textbf{Model}
& \multicolumn{3}{c}{\textbf{V$^{*}$Bench}}
& \multicolumn{3}{c}{\textbf{HRBench-4K}}
& \multicolumn{3}{c}{\textbf{HRBench-8K}} \\
\cmidrule(lr){2-4} \cmidrule(lr){5-7} \cmidrule(lr){8-10}
& Attribute & Spatial & Overall
& Single & Cross & Overall
& Single & Cross & Overall \\
\midrule
LLaVA-OV-7B & 78.3 & 65.8 & 73.3 & 73.0 & 54.5 & 63.8 & 65.0 & 52.0 & 58.5 \\
\rowcolor{gray!12}
\quad w/ \textbf{VisLens}$_{\Delta}$
& 87.0$_{\textcolor[HTML]{2E8B57}{8.7}}$ & 73.7$_{\textcolor[HTML]{2E8B57}{7.9}}$ & 81.7$_{\textcolor[HTML]{2E8B57}{8.4}}$
& 82.0$_{\textcolor[HTML]{2E8B57}{9.0}}$ & 56.8$_{\textcolor[HTML]{2E8B57}{2.3}}$ & 69.4$_{\textcolor[HTML]{2E8B57}{5.6}}$
& 77.8$_{\textcolor[HTML]{2E8B57}{12.8}}$ & 59.5$_{\textcolor[HTML]{2E8B57}{7.5}}$ & 68.6$_{\textcolor[HTML]{2E8B57}{10.1}}$ \\
Qwen2.5-VL-7B & 67.0 & 61.8 & 64.9 & 71.5 & 52.5 & 62.1 & 62.5 & 50.5 & 56.5 \\
\rowcolor{gray!12}
\quad w/ \textbf{VisLens}$_{\Delta}$
& 76.5$_{\textcolor[HTML]{2E8B57}{9.5}}$ & 71.1$_{\textcolor[HTML]{2E8B57}{9.3}}$ & 74.3$_{\textcolor[HTML]{2E8B57}{9.4}}$
& 81.0$_{\textcolor[HTML]{2E8B57}{9.5}}$ & 60.2$_{\textcolor[HTML]{2E8B57}{7.7}}$ & 70.6$_{\textcolor[HTML]{2E8B57}{8.5}}$
& 74.5$_{\textcolor[HTML]{2E8B57}{12.0}}$ & 55.8$_{\textcolor[HTML]{2E8B57}{5.3}}$ & 65.1$_{\textcolor[HTML]{2E8B57}{8.6}}$ \\
InternVL3-8B & 70.4 & 73.7 & 71.7 & 80.5 & 59.2 & 69.9 & 68.2 & 54.8 & 61.5 \\
\rowcolor{gray!12}
\quad w/ \textbf{VisLens}$_{\Delta}$
& 80.9$_{\textcolor[HTML]{2E8B57}{10.5}}$ & 82.9$_{\textcolor[HTML]{2E8B57}{9.2}}$ & 81.7$_{\textcolor[HTML]{2E8B57}{10.0}}$
& 89.0$_{\textcolor[HTML]{2E8B57}{8.5}}$ & 63.0$_{\textcolor[HTML]{2E8B57}{3.8}}$ & 76.0$_{\textcolor[HTML]{2E8B57}{6.1}}$
& 79.2$_{\textcolor[HTML]{2E8B57}{11.0}}$ & 56.5$_{\textcolor[HTML]{2E8B57}{1.7}}$ & 67.9$_{\textcolor[HTML]{2E8B57}{6.4}}$ \\
\bottomrule
\end{tabular}%
}
\end{table*}

\begin{table}[htbp]
\centering
\caption{Performance on Visual-Search QA benchmarks. Subscripts in
\textcolor[HTML]{2E8B57}{green} denote the improvement of \textbf{VisLens}
over each base model.}
\label{tab:visual-search-qa}
\small
\setlength{\tabcolsep}{6pt}
\renewcommand{\arraystretch}{1.05}
\begin{tabular}{l ccc}
\toprule
\textbf{Model} & \textbf{V$^{*}$Bench} & \textbf{HRBench-4K} & \textbf{HRBench-8K} \\
\midrule
LLaVA-OV-7B & 73.3 & 63.8 & 58.5 \\
\rowcolor{gray!12}
\quad w/ \textbf{VisLens}$_{\Delta}$
& 81.7$_{\textcolor[HTML]{2E8B57}{8.4}}$
& 69.4$_{\textcolor[HTML]{2E8B57}{5.6}}$
& 68.6$_{\textcolor[HTML]{2E8B57}{10.1}}$ \\
Qwen2.5-VL-7B & 64.9 & 62.1 & 56.5 \\
\rowcolor{gray!12}
\quad w/ \textbf{VisLens}$_{\Delta}$
& 74.3$_{\textcolor[HTML]{2E8B57}{9.4}}$
& 70.6$_{\textcolor[HTML]{2E8B57}{8.5}}$
& 65.1$_{\textcolor[HTML]{2E8B57}{8.6}}$ \\
InternVL3-8B & 71.7 & 69.9 & 61.5 \\
\rowcolor{gray!12}
\quad w/ \textbf{VisLens}$_{\Delta}$
& 81.7$_{\textcolor[HTML]{2E8B57}{10.0}}$
& 76.0$_{\textcolor[HTML]{2E8B57}{6.1}}$
& 67.9$_{\textcolor[HTML]{2E8B57}{6.4}}$ \\
\bottomrule
\end{tabular}
\end{table}

Tab.~\ref{tab:visual-search-qa} reports results on visual-search QA benchmarks
across three MLLM backbones. VisLens improves every backbone on every benchmark,
showing that the gains are not tied to a specific model architecture. The largest
improvements appear in the single-object settings, namely the Attribute
split of V$^{*}$Bench and the Single split of HR-Bench. These questions require
the model to identify a small localized target before answering, so they directly
test the localization bottleneck that VisLens is designed to address. In these
settings, VisLens yields substantial gains, including +10.5 points on Direct
Attribute for InternVL3 and +12.8/+12.0/+11.0 points on HR-Bench-8K Single for
LLaVA-OneVision, Qwen2.5-VL, and InternVL3 respectively.

VisLens also improves relational questions, including the Spatial split of
V$^{*}$Bench and the Cross split of HR-Bench. These gains are positive across all
backbones, but are generally smaller than those in the single-object setting. This
is expected, since relational questions require not only sharper object perception, but
also enough surrounding context to compare multiple targets. Cropping can make
each target more legible, but overly tight crops may remove part of the spatial
context needed for relation reasoning. The consistent gains therefore suggest
that VisLens improves object-level perception while retaining enough global
context, through the original image, to support many relational queries.

\subsection{Accuracy--Latency Trade-off}

\begin{figure*}[htbp]
  \centering
  \includegraphics[width=0.33\textwidth]{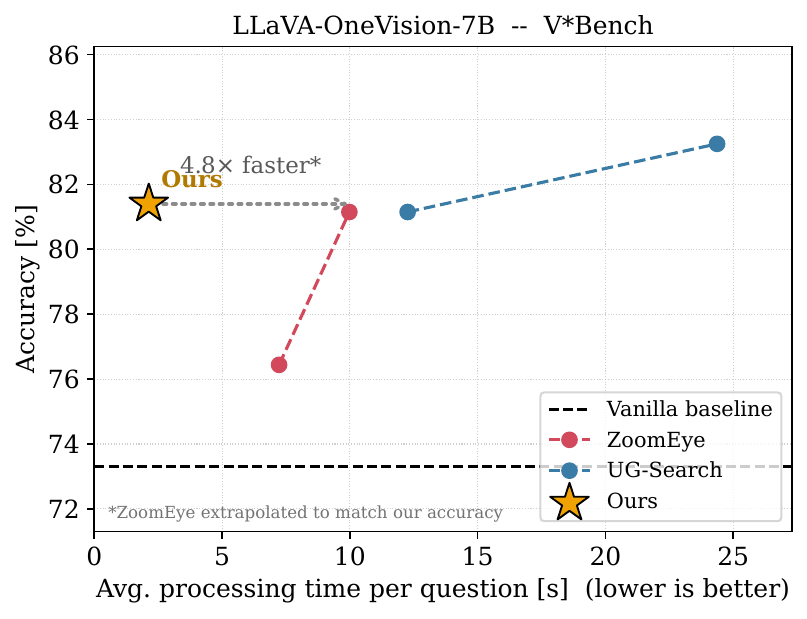}\hfill
  \includegraphics[width=0.33\textwidth]{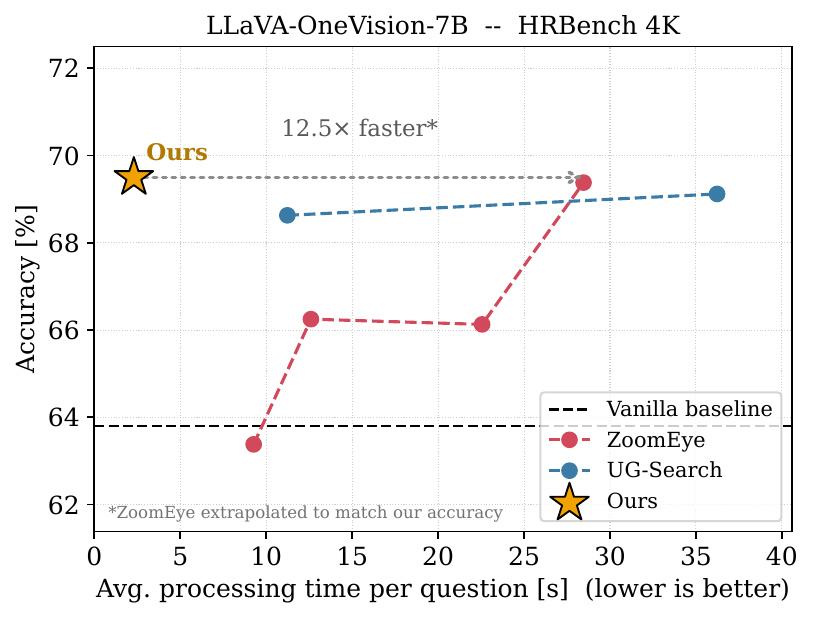}\hfill
  \includegraphics[width=0.33\textwidth]{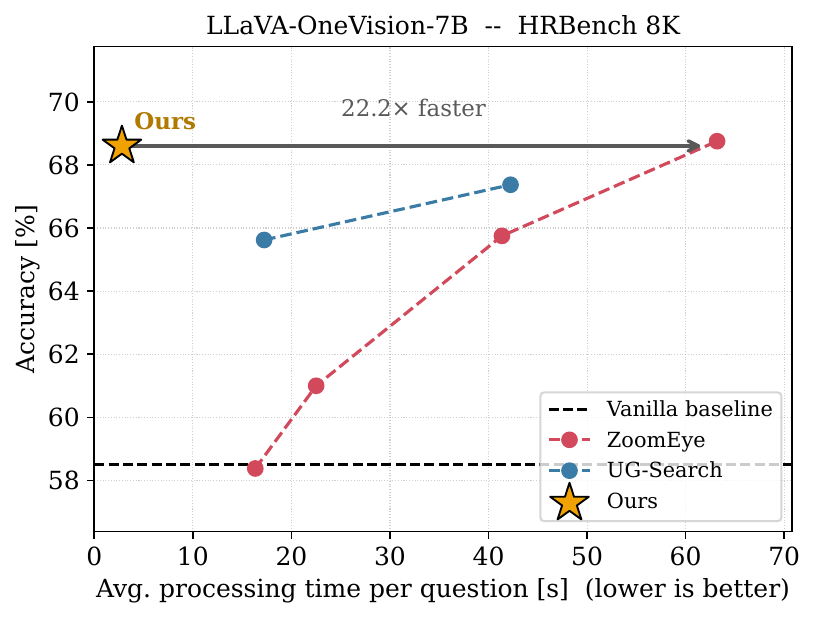}
  \caption{Accuracy--latency trade-off on V$^{*}$Bench, HRBench-4K, and
  HRBench-8K (LLaVA-OV-7B backbone). Accuracy is plotted against the average
  processing time in seconds per question. Our method lies on the
  Pareto frontier, on par with or above ZoomEye and other multi-pass baselines
  at a fraction of the cost.}
  \label{fig:acc-latency}
\end{figure*}
\begin{table}[htbp]
\centering
\scriptsize
\caption{Comparison with Thyme\cite{zhang2025thyme} (Qwen2.5-VL-7B backbone). Accuracy in \%, runtime in seconds per example; ↑/↓ indicate higher/lower is better.}
\label{tab:qwen_latency}
\begin{tabular}{lccccc}
\toprule
\multirow{2}{*}{Benchmark}
& \multicolumn{2}{c}{Accuracy $\uparrow$}
& \multicolumn{2}{c}{Time (s) $\downarrow$}
& \multirow{2}{*}{Speed-up $\uparrow$} \\
\cmidrule(lr){2-3}
\cmidrule(lr){4-5}
& Thyme\cite{zhang2025thyme} & Ours
& Thyme\cite{zhang2025thyme} & Ours
& \\
\midrule
V$^{*}$Bench
& 72.25 & \textbf{74.30}
& 8.78 & \textbf{0.98}
& \textbf{9.0$\times$} \\
HRBench-4K
& 69.87 & \textbf{70.60}
& 10.30 & \textbf{1.04}
& \textbf{9.9$\times$} \\
HRBench-8K
& \textbf{65.12} & 65.10
& 12.51 & \textbf{1.47}
& \textbf{8.5$\times$} \\
\bottomrule
\end{tabular}
\end{table}
Tab.~\ref{tab:qwen_latency} makes the accuracy–latency trade-off explicit. Thyme reaches its accuracy through a supervised fine-tuning (SFT) and RL pipeline that must be trained on a purpose-built corpus of code-interleaved reasoning trajectories. At inference it runs a multi-turn loop, repeatedly emitting code, invoking an external sandbox, and re-reasoning over the returned image. Our method needs neither the trajectory-construction pipeline nor the RL stage: a single forward pass matches or exceeds Thyme's accuracy on V*Bench, HRBench-4K, and HRBench-8K while cutting latency by 8.5–9.9×.

The same trade-off holds against multi-pass search baselines on a second backbone. Fig.~\ref{fig:acc-latency} plots accuracy versus per-question latency for ZoomEye\cite{shen2024zoomeye} and UG-Search\cite{kim2025training} on LLaVA-OV-7B; both rely on iterative zoom or search passes and so trace out a latency-for-accuracy curve. Our method sits at the top-left of every panel, on the Pareto frontier, staying on or above these baselines' curve in a single pass. Relative to ZoomEye\cite{shen2024zoomeye} extrapolated to our accuracy, this is 4.8$\times$, 12.5$\times$, and 22.2$\times$ faster on V*Bench, HRBench-4K, and HRBench-8K respectively. Notably the gap widens with image resolution, precisely where multi-pass methods scale worst.

\subsection{Effect on General VQA}
\begin{table}[!ht]
  \centering
  \scriptsize
  \caption{Standard QA accuracy (\%). Subscripts give the change of
  \textbf{VisLens} relative to its frozen backbone:
  \textcolor[HTML]{2E8B57}{green} marks a gain,
  \textcolor[HTML]{B22222}{red} a drop.}
  \label{tab:standard-qa}
  \setlength{\tabcolsep}{6pt}
  \renewcommand{\arraystretch}{0.95}
\begin{tabular}{lccc}
\toprule
\textbf{Model} & A-OKVQA & POPE & GQA \\
\midrule
LLaVA-OV-7B   & 90.6 & 89.2 & 62.7 \\
\rowcolor{gray!12}
\quad w/ \textbf{VisLens}
& 90.8$_{\textcolor[HTML]{2E8B57}{0.2}}$
& 90.2$_{\textcolor[HTML]{2E8B57}{1.0}}$
& 62.9$_{\textcolor[HTML]{2E8B57}{0.2}}$ \\
Qwen2.5-VL-7B & 87.6 & 87.6 & 60.7 \\
\rowcolor{gray!12}
\quad w/ \textbf{VisLens}
& 88.5$_{\textcolor[HTML]{2E8B57}{0.9}}$
& 87.2$_{\textcolor[HTML]{B22222}{0.4}}$
& 59.9$_{\textcolor[HTML]{B22222}{0.8}}$ \\
InternVL3-8B  & 88.7 & 90.8 & 62.3 \\
\rowcolor{gray!12}
\quad w/ \textbf{VisLens}
& 88.3$_{\textcolor[HTML]{B22222}{0.4}}$
& 90.8$_{\textcolor{black!55}{0.0}}$
& 63.2$_{\textcolor[HTML]{2E8B57}{0.9}}$ \\
\bottomrule
\end{tabular}
\end{table}

\noindent\textbf{No degradation on standard VQA.} Tab. ~\ref{tab:standard-qa}
confirms that VisLens preserves performance on regular-size images that lack a
small-target bottleneck. Across A-OKVQA, POPE, and GQA, VisLens stays within
about one point of each frozen baseline, so VisLens can be left always on,
enhancing fine-grained perception where it matters without harming general VQA.

\subsection{Ablation: Source Layer for the Tuned-Lens}
\begin{table}[htbp]
  \centering
  \scriptsize
  \setlength{\tabcolsep}{10pt}
\caption{Source-layer ablation for the tuned-lens. The final-layer baseline decodes the MLLM's actual final-layer visual-token states. Higher is better.}
\label{tab:src-depth-ablation}
\begin{tabular}{lccc}
\toprule
\textbf{Source layer} &Token detection & Crop inclusion & GT coverage \\
\midrule
Pre-layer 1 & 0.567 & \textbf{0.390} & \textbf{0.641} \\
Layer 1     & \textbf{0.571} & 0.385 & 0.635 \\
Layer 8     & 0.562 & 0.387 & 0.636 \\
Layer 15    & 0.569 & 0.384 & 0.631 \\
Layer 22    & 0.570 & 0.386 & 0.632 \\
\midrule
Baseline    & 0.582 & 0.393 & 0.653 \\
\bottomrule
\end{tabular}
\end{table}
\noindent\textbf{Setup.} 
We evaluate where the tuned-lens should read from by sweeping the source layer on
a held-out subset of small COCO~\cite{coco} objects. The source
layer determines how early VisLens can stop the backbone before decoding
visual-token semantics. We compare source layers from the post-projector
representation, denoted pre-layer 1, to several later language-model layers. We
also report a final-layer baseline, which applies the same decoding and matching
pipeline to the MLLM's actual final-layer visual-token states after a full
forward pass. This is the teacher readout used to train the tuned-lens and serves
as an upper reference for the translated early-layer readouts. We report three
localization metrics: \emph{token detection}, the fraction of objects whose
target word or synonym is decoded in an overlapping visual cell; \emph{crop
inclusion}, the fraction whose constructed crop fully contains the ground-truth
box; and \emph{GT coverage}, the average fraction of the ground-truth box covered
by the crop.

\noindent\textbf{Results.}
Tab.~\ref{tab:src-depth-ablation} shows that localization quality is nearly flat
across source layers: the spread among translated readouts is at most
0.010 absolute per metric, with no monotonic trend in depth. The
final-layer baseline, which requires a full forward pass, remains the
strongest readout, but only by a small margin. In other words, stopping the backbone at the earliest
possible point costs at most $\sim$0.02 absolute localization quality
relative to running the model to completion. Notably, the pre-layer 1
representation, taken directly after the visual projector before any
language-model layer, is already competitive with all deeper source layers.

\noindent\textbf{Implication.}
This result supports the central design choice of VisLens: the information needed
for small-object localization is already present in very early visual-token states.
The tuned-lens does not need to wait for late language-model layers to obtain a
usable semantic map. We therefore use pre-layer 1 as the default source layer.
\subsection{Ablation: Crop Construction}

\noindent\textbf{Separate versus union crops.}
Tab.s~\ref{tab:ablate-union} and~\ref{tab:ablation-crop} compare two ways of
constructing the final crop input. VisLens-sep keeps one crop per target, whereas
VisLens-union merges all matched targets into a single crop. The better choice
depends on image resolution and target spread. On V$^*$Bench, VisLens-union
performs best, because the relevant targets are often close enough that one crop
can preserve their spatial relation without adding much background. On HR-Bench,
especially at 8K resolution, VisLens-sep becomes stronger. In this regime, a
single union crop can span distant regions, include large amounts of background,
and shrink each target after resizing. Separate crops preserve more local detail
and suppress distractors.

\noindent\textbf{Crop hyperparameters.}
The same pattern appears in the crop-level ablations. A small link distance works
best, showing that only nearby components should be merged. The area cap
$\tau=0.2$ gives the best balance between context and distractors: smaller crops
can miss useful surrounding evidence, while larger crops dilute the target again.
Minimal cell padding is also sufficient, suggesting that VisLens benefits most
from tight object-centered crops rather than broad scene crops.
\begin{table}[htbp]
\centering
\caption{Merging ablation (Qwen2.5-VL-7B). sep = one crop per target; union = merged.}
\label{tab:ablate-union}
\scriptsize
\setlength{\tabcolsep}{6pt}
\renewcommand{\arraystretch}{0.95}
\begin{tabular}{lccc}
\toprule
\textbf{Model} & V$^{*}$Bench & HRBench-4K & HRBench-8K \\
\midrule
Qwen2.5-VL-7B & 64.9 & 62.1 & 56.5 \\
\rowcolor{gray!12}
\quad w/ \textbf{VisLens-sep}$_{\Delta}$
& 74.3$_{\textcolor[HTML]{2E8B57}{9.4}}$
& 70.6$_{\textcolor[HTML]{2E8B57}{8.5}}$
& 65.1$_{\textcolor[HTML]{2E8B57}{8.6}}$ \\
\rowcolor{gray!12}
\quad w/ \textbf{VisLens-union}$_{\Delta}$
& 75.9$_{\textcolor[HTML]{2E8B57}{11.0}}$
& 71.0$_{\textcolor[HTML]{2E8B57}{8.9}}$
& 62.0$_{\textcolor[HTML]{2E8B57}{5.5}}$ \\
\bottomrule
\end{tabular}
\end{table}

\begin{table}[htbp]
\centering
\scriptsize
\caption{Crop-hyperparameter ablation on V$^{*}$Bench (Qwen2.5-VL-7B).
Numbers are overall accuracy (\%); higher is better.
$^{\ast}$ marks the default setting.}
\label{tab:ablation-crop}
\begin{tabular}{c@{\hskip 8pt}c @{\hskip 20pt} c@{\hskip 8pt}c @{\hskip 20pt} c@{\hskip 8pt}c}
\toprule
\multicolumn{2}{c}{\textbf{Area cap}} & \multicolumn{2}{c}{\textbf{Cell padding}} & \multicolumn{2}{c}{\textbf{Link distance}} \\
\cmidrule(lr){1-2} \cmidrule(lr){3-4} \cmidrule(lr){5-6}
$\tau$ & Acc. & cells & Acc. & gap & Acc. \\
\midrule
0.1          & 68.6          & 1$^{\ast}$ & \textbf{74.3} & 1$^{\ast}$ & \textbf{74.3} \\
0.2$^{\ast}$ & \textbf{74.3} & 2          & 73.3          & 2          & 72.3 \\
0.3          & 73.3          & 3          & 73.8          & 3          & 73.3 \\
0.4          & 73.8          & 4          & 73.8          & 4          & 73.3 \\
\bottomrule
\end{tabular}
\end{table}

\subsection{Ablation: Replacing the Lens Matcher with an External Detector}

\noindent\textbf{Setup.}
To test whether the gains come simply from adding a crop before answering, we
replace the VisLens matcher with an external open-vocabulary segmenter, SAM 3\cite{sam3},
while keeping the rest of the pipeline unchanged. For each extracted target, SAM 3 is prompted with the target name, and the highest-confidence mask above the
default threshold is converted into a crop.

\noindent\textbf{Results.}
Tab.~\ref{tab:vstar-sam} shows that SAM 3 improves over the vanilla backbone, but the
gains are much smaller than those of VisLens. The main failure mode is recall:
SAM 3 often fails to return a valid crop for small or visually ambiguous targets.
In contrast, VisLens produces crops for more than 90\% of queries because it does
not require an external detector to recognize the object. Instead, it reads the
object evidence already encoded in the MLLM's own visual-token representations.

\noindent\textbf{Implication.}
This ablation shows that VisLens is not only a crop-and-reprompt heuristic. Its
advantage comes from the lens-based matcher, which exposes internal semantic
evidence that may be unavailable to a standalone detector.
\begin{table}[htbp]
\centering
\caption{Performance on V$^*$Bench using SAM3 for detection, reported at SAM3 default confidence score 0.5.}
\label{tab:vstar-sam}
\small
\setlength{\tabcolsep}{6pt}
\renewcommand{\arraystretch}{0.95}
\begin{tabular}{lccc}
\toprule
\textbf{V$^*$Bench} & LLaVA-OV-7B & InternVL3-8B & Qwen2.5-VL-7B \\
\midrule
Baseline & 73.3 & 71.7 & 64.9 \\
\rowcolor{gray!12}
w/ \textbf{SAM 3}$_{\Delta}$
& 74.3$_{\textcolor[HTML]{2E8B57}{1.0}}$
& 77.0$_{\textcolor[HTML]{2E8B57}{5.3}}$
& 71.2$_{\textcolor[HTML]{2E8B57}{6.3}}$ \\
\rowcolor{gray!12}
w/ \textbf{VisLens}$_{\Delta}$
& 81.7$_{\textcolor[HTML]{2E8B57}{8.4}}$
& 81.7$_{\textcolor[HTML]{2E8B57}{10.0}}$
& 74.3$_{\textcolor[HTML]{2E8B57}{9.4}}$ \\
\bottomrule
\end{tabular}
\end{table}

\section{Analysis}

\begin{figure}[htbp]
  \centering
  \includegraphics[clip, trim=30 290 25 20, width=\linewidth]{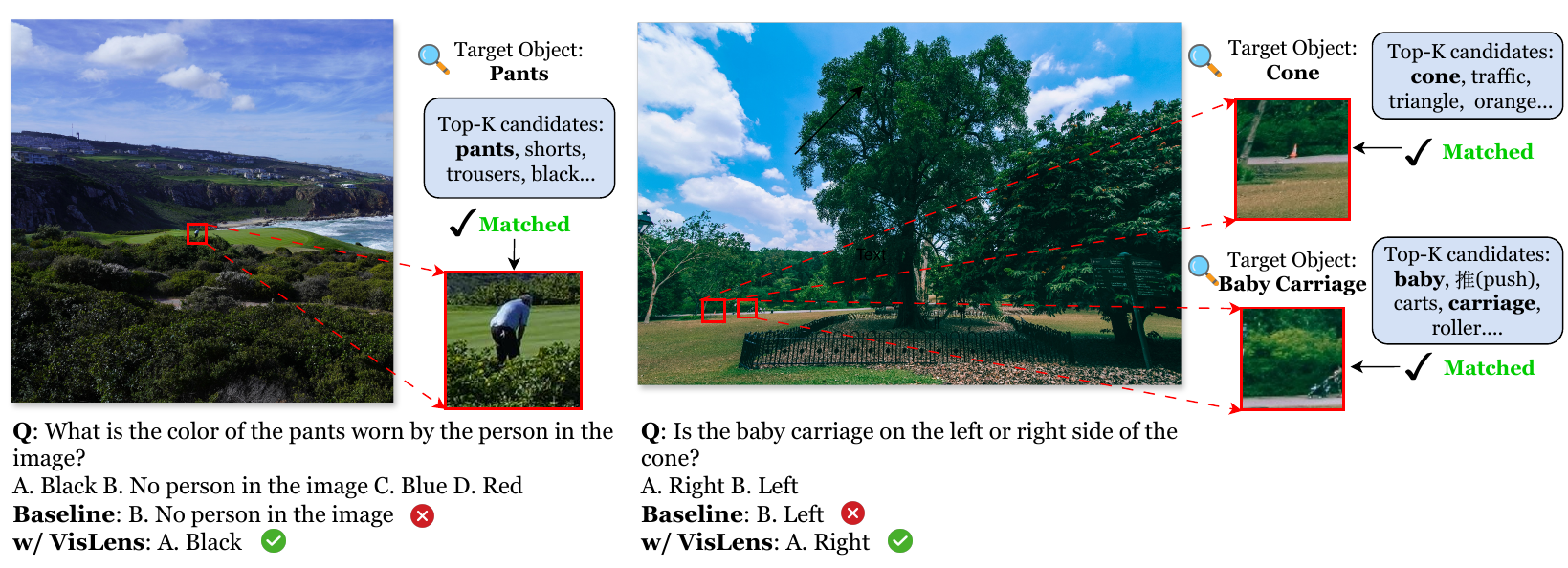}
  \caption{Qualitative results. Extracted targets, tuned-lens top-K tokens, and answers with/without VisLens.}
  \label{fig:qualitative}
\end{figure}

\subsection{When Local Semantics Are Present but Underused}
\label{sec:qualitative}

Fig.~\ref{fig:qualitative} illustrates how VisLens uses vocabulary-level
semantic evidence recovered from translated early visual-token states. The tuned-lens maps an early hidden state into the final-layer decoding space, producing a
local readout for each visual cell. VisLens then turns this readout into an
explicit crop and re-prompts the frozen model with the original image and the
localized region.

\noindent\textbf{Single-object attribute.}
In the left example, the question asks for the color of a person's pants, while
the baseline answers that no person is present. The tuned-lens readout around the
relevant region contains \texttt{pants}, \texttt{shorts}, \texttt{trousers}, and
\texttt{black}. The readout indicates that object-level semantics are locally recoverable from the visual-token representation. By cropping this region and feeding it back with the original image, VisLens makes the relevant evidence more salient for answer generation.

\noindent\textbf{Multiple targets and relational reasoning.}
The right example asks for the position of a baby carriage relative to a cone.
Here, VisLens recovers semantic evidence for both targets: the carriage region is
associated with tokens such as \texttt{baby}, \texttt{cart}, and
\texttt{carriage}, while the cone region is associated with \texttt{cone},
\texttt{traffic}, and \texttt{triangle}. The method therefore does not create new
visual information; it converts locally readable semantics into target crops. The
original image is kept in the prompt so that the model can still use global
spatial context when answering the relational question.
\begin{figure}[htbp]
  \centering
  \includegraphics[clip, trim=30 270 25 20, width=\linewidth]{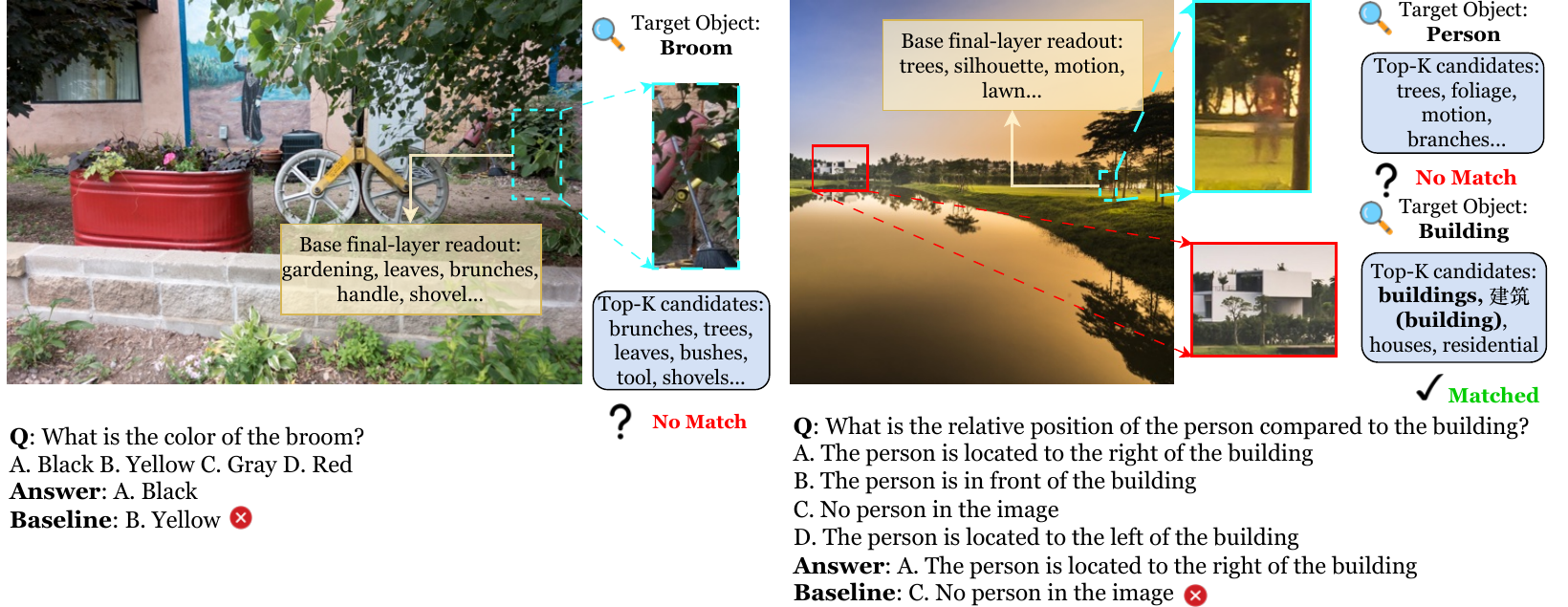}
    \caption{Interpretation of failures. Tuned-lens and base-model top logits for two missed targets.}
  \label{fig:interpretation}
\end{figure}

\subsection{Interpreting Failures Through Decoded Tokens}
\label{sec:interp}

The decoded tokens used by VisLens also make its failures inspectable. When a
crop is not produced, the local readout shows which semantic evidence was
available to the matcher. In this sense, a VisLens failure is not only an answer failure, but also a readable localization failure: the target is either absent from the vocabulary-level readout, expressed through a related but non-matching concept, or dominated by nearby scene context.

Fig.~\ref{fig:interpretation} shows two examples. In the broom case, the target
is visible in the image, but the tuned-lens readout around the relevant region is
dominated by nearby scene and tool concepts such as \texttt{branches},
\texttt{trees}, \texttt{leaves}, and \texttt{shovel}. The base model's
final-layer readout shows a similar pattern, with tokens such as
\texttt{gardening}, \texttt{leaves}, \texttt{branches}, \texttt{handle}, and
\texttt{shovel}. The two readouts therefore point to the same source of
difficulty: the region is represented through semantically related context, but
not through a token that safely matches the queried object \texttt{broom}.

The person--building case shows an asymmetric failure. The building is
semantically accessible and is matched, while the person region is decoded mainly
as background-like content. The tuned-lens readout contains tokens such as
\texttt{trees}, \texttt{foliage}, \texttt{motion}, and \texttt{branches}, and the
base model's final-layer readout similarly emphasizes \texttt{trees},
\texttt{silhouette}, \texttt{motion}, and \texttt{lawn}. The missed target is
therefore not random: both readouts describe visual evidence near the person, but
they do not express it as the target category \texttt{person}. As a result,
VisLens recovers the building but not the person, so the relational query remains
unresolved.

These examples show how VisLens turns failures into interpretable evidence. The
failure mode belongs to the lens-based matching pipeline, yet it often follows
the same semantic pattern visible in the base model's final-layer readout. This
connection helps explain why the crop was not formed: the target is not reliably
available as a matching vocabulary-level concept, even though nearby or related
visual evidence may still be present in the readout.

\subsection{Limitations and Remaining Challenges}
\label{sec:limitations}

We close the analysis by summarizing recurring cases where the current design is
less effective. VisLens is built around local vocabulary-level semantics: each
visual cell is decoded into tokens, and crops are formed when these tokens match
the query target. This works well for object-centric visual search, but is less
reliable when the relevant evidence is symbolic, textual, or distributed across
global structure.

Diagram-like images are a typical example. In high-resolution benchmarks, a
diagram may be split into many small patches, where each patch contains only a
fragment of the relevant content, such as a single digit, a letter, an arrow
segment, or part of a line. The tuned-lens readout for such a patch may therefore
produce tokens for isolated local elements rather than the higher-level concept
described in the question. Since no individual patch exposes a reliable semantic
match to the target, the matcher may not localize the intended region.

This limitation follows from the local nature of the current matching step.
VisLens is most effective when the target corresponds to a visually localizable
object or object part whose semantics can be recovered from nearby visual tokens.
It is less suited to tasks requiring OCR, mathematical notation, chart
understanding, or reasoning over the global layout of a diagram. Extending the
method to these settings would require structural grouping or relation-aware
matching beyond independent per-patch semantics.
\section{Conclusion}

We introduced VisLens, a single-pass and interpretable Visual Search method for
MLLMs. VisLens reads local visual-token semantics through a tuned-lens, matches
decoded tokens to query targets, and turns the matched regions into explicit
crops for crop-guided answering. The base MLLM remains frozen; only a lightweight
translator is trained.

Across V$^*$Bench and HR-Bench, VisLens consistently improves visual-search QA,
with the largest gains on single-object questions where small-target localization
is the main bottleneck. It also sits on the accuracy--latency Pareto frontier across all evaluated benchmarks, with the advantage over multi-pass search widening as image
resolution grows, which is precisely the regime visual search targets.

VisLens also makes localization decisions inspectable: each crop is grounded in
decoded vocabulary tokens, and the same tokens reveal when target evidence is
locally recoverable or when the matcher instead follows related surrounding
semantics. The current design is strongest for object-centric visual search, and
less suited to symbolic, textual, chart-like, or diagrammatic inputs where
evidence is distributed across structure. Extending lens-based localization with
OCR, structural grouping, and relation-aware matching is a direction for future
work.

\section*{Acknowledgements.}
The authors gratefully acknowledge the scientific support and resources of the AI service infrastructure \textit{LRZ AI Systems} provided by the Leibniz Supercomputing Centre (LRZ) of the Bavarian Academy of Sciences and Humanities (BAdW), funded by Bayerisches Staatsministerium für Wissenschaft und Kunst (StMWK).

\bibliographystyle{splncs04}
\bibliography{main}

\end{document}